\documentclass[10pt, a4paper]{article}

\usepackage{lrec-coling2024} 
\usepackage{multirow}
\usepackage{amsmath}
\title{Deciphering Political Entity Sentiment in News with Large Language Models: Zero-Shot and Few-Shot Strategies}

\name{Alapan Kuila,  Sudeshna Sarkar} 

\address{IIT Kharagpur \\
         India \\
         alapan.cse@iitkgp.ac.in, sudeshna@cse.iitkgp.ac.in}

\abstract{
Sentiment analysis plays a pivotal role in understanding public opinion, particularly in the political domain where the portrayal of entities in news articles influences public perception. In this paper, we investigate the effectiveness of Large Language Models (LLMs) in predicting entity-specific sentiment from political news articles. Leveraging zero-shot and few-shot strategies, we explore the capability of LLMs to discern sentiment towards political entities in news content. Employing a chain-of-thought (COT) approach augmented with rationale in few-shot in-context learning, we assess whether this method enhances sentiment prediction accuracy. Our evaluation on sentiment-labeled datasets demonstrates that LLMs, outperform fine-tuned BERT models in capturing entity-specific sentiment. We find that learning in-context significantly improves model performance, while the self-consistency mechanism enhances consistency in sentiment prediction. Despite the promising results, we observe inconsistencies in the effectiveness of the COT prompting method. Overall, our findings underscore the potential of LLMs in entity-centric sentiment analysis within the political news domain and highlight the importance of suitable prompting strategies and model architectures.
 \\ \newline  }

\begin{document}

\maketitleabstract

\section{Introduction}

Sentiment analysis (SA) is a vital area in natural language processing (NLP)\cite{liu2020sentiment}, focused on deciphering opinions and emotions using computational methods~\cite{Poria2020BeneathTT}. It has diverse applications, from product reviews to social media insights. Previous research has addressed sentiment analysis at various levels, such as sentence, paragraph, and document levels~\cite{zhang2023sentiment}. Moreover, studies have focused on different targets of sentiment, including overall sentiment, aspect-based sentiment~\cite{brun2018aspect}, and sentiment associated with event mentions~\cite{zhang2022enhancing}. Analyzing sentiment pertinent to the salient entities in the news article is an important problem in computational journalism and news content analysis~\cite{ronningstad2023entity}. In the context of political natural language processing (NLP), understanding the sentiment towards political entities in news articles is particularly crucial. Political entities, such as countries, politicians, and political organizations, often drive the narrative in news coverage. Therefore, being able to accurately assess the sentiment towards these entities can provide valuable insights into public opinion, political discourse, and media framing.

Similar to works by \cite{tang-etal-2023-finentity}, and \cite{bastan2020author}, our research specifically targets sentiment analysis related to particular entities. Historically, sentiment analysis relied on bag-of-word models, which failed to capture word ordering, a crucial aspect of sentiment prediction. Later, machine learning (ML) and deep learning (DL) models gained popularity for sentiment analysis tasks, though they struggled with generalization on domain-specific datasets~\cite{kenyon-dean-etal-2018-sentiment}. Recently, techniques like transfer learning~\cite{golovanov-etal-2019-large} and self-supervised learning~\cite{qian-etal-2023-sentiment} have been applied to improve model generalization and reduce data dependence, particularly demonstrating promising performance in few-shot settings with limited annotated data. However, state-of-the-art deep neural network models remain complex and opaque in their decision-making processes, posing challenges for both end-users and system designers.

However, recent research on pre-trained large language models (LLMs) has demonstrated impressive performance across a variety of natural language processing (NLP) tasks, particularly in common sense reasoning~\cite{brown2020language}. These LLMs have proven capable of generalizing to new tasks using zero-shot and few-shot learning, facilitated by suitable prompts and in-context learning~\cite{huang2022towards}. Moreover, the introduction of chain-of-thought (COT) prompting~\cite{wei2022chain} has further enhanced the reasoning abilities of LLMs by generating intermediate reasoning steps. By incorporating rationale into the prompt design and providing (input, output) instance-pair demonstrations, the COT approach encourages LLMs to generate textual explanations alongside predicting the final output. Additionally, self-consistency mechanisms~\cite{Wang2022SelfConsistencyIC} reinforce the reasoning capabilities of LLMs through \textit{sample-and-marginalize} decoding procedures. Despite these advancements, some studies have suggested that accumulating explanations with prompts during in-context learning may have adverse effects on LLM performance in question-answering (QA) and natural language inference (NLI) tasks~\cite{Ye2022TheUO}. Nevertheless, LLMs have proven effective in various textual reasoning tasks, including arithmetic and symbolic reasoning problems~\cite{Wang2022SelfConsistencyIC}. In our research, we aim to investigate whether LLMs can accurately predict entity-centric sentiment polarity from political news text. By exploring the intersection of large language models and sentiment analysis, we hope to shed light on the capabilities and limitations of these models in understanding and interpreting sentiment dynamics in textual data.

To employ large language models (LLMs) for predicting entity-specific sentiment, we harness the chain-of-thought (COT) mechanism to guide prompt design. In our zero-shot chain-of-thought approach, we adopt a two-stage prompting strategy. Initially, we extract the contextual justification for the prediction, followed by returning the final sentiment label in the second stage. Our few-shot approach involves integrating a limited number of (entity context, entity-centric sentiment label, rationale) triplets into the LLMs during training. Here, the entity context may encompass a sentence, paragraph, or entire document, while the entity-centric sentiment label denotes the sentiment polarity towards the target entity as depicted in the context. The rationale comprises one or more sentences elucidating the reasoning behind the predicted outcome. We assess our model's performance based on the accuracy of the final predicted sentiment class.

Additionally, previous research has highlighted the necessity of scaling up LLMs with several hundred billion parameters, such as the OpenAI GPT series (GPT-3-175B)~\cite{brown2020language}, PaLM(540B)~\cite{Chowdhery2022PaLMSL}, and LaMDA(137B)~\cite{Thoppilan2022LaMDALM}, to achieve satisfactory performance in COT scenarios. However, adopting these large pre-trained language models may be unfeasible for many users with resource constraints~\cite{ranaldi-freitas-2024-aligning}. In our experiments, we employ LLMs with relatively fewer model parameters, namely Mistral-7B~\cite{jiang2023mistral}, LLaMA2-13B~\cite{touvron2023llama} and Falcon-40B~\cite{Almazrouei2023TheFS}. By deliberately altering various aspects of the demonstrated rationale and conducting a series of ablation experiments, we measure how the model's performance varies accordingly. Our extensive results demonstrate the effectiveness of LLMs with relatively fewer parameters in the task of entity-centric sentiment prediction from political news articles.

Our contributions are as follows:
\begin{itemize}

    \item We explore the capability of LLMs to predict entity-specific sentiment from the news context in a zero-shot setting.
    
    \item We examine the efficacy of the Chain-of-Thought (COT) approach in conjunction with Large Language Models (LLMs), bolstered by rationale in few-shot in-context learning. Our objective is to determine whether this combined approach improves the model's ability to predict entity-specific sentiment from document-level context.

    \item We evaluate the accuracy and robustness of our proposed approach using two sentiment-labeled news datasets. The first dataset~\footnote{\url{https://github.com/alapanju/EntSent}} comprises political news articles sourced from the Event-Registry API~\footnote{\url{https://github.com/EventRegistry/event-registry-python}}, while the second dataset~\footnote{\url{https://github.com/StonyBrookNLP/PerSenT}} is obtained from \cite{bastan2020author}, providing diverse contexts for evaluation and comparison.

\end{itemize}

\section{Entity Centric Sentiment from Political News}
\label{sec:append-how-prod}

In this paper, we address the task of determining the overall sentiment polarity expressed towards a target entity in a political news article. This task differs significantly from existing works on sentiment prediction in movie reviews, product reviews, or social media datasets~\cite{KUMARESAN2023100663}. Unlike review text or social media datasets, news articles contain descriptive content with a significant amount of redundant information that is often irrelevant for sentiment prediction of the target entity. Additionally, subjective opinions are sometimes presented as objective information, posing challenges for automatic classifiers. While it may be easy for humans to discern the inherent sentiment polarity, automatic classifiers face difficulties in extracting the target entity-specific context from news articles, especially when multiple entities are mentioned multiple times, both directly and indirectly~\cite{Fei2023ReasoningIS}.

Moreover, a single news article may contain multiple opinions directed towards the target entity, and the sentiment towards the same entity may vary across different paragraphs within the same article. Hence, navigating through irrelevant information to extract the target entity-specific context and predict the correct sentiment becomes challenging.

In the following sections, we first describe our approach, followed by the experimental details, including the dataset used and the LLM models employed. Subsequently, we present our experimental findings and engage in pertinent discussions. Following this, we delve into a detailed examination of existing works within this domain. Finally, the paper concludes by summarizing key discoveries and outlining avenues for future research.

\section{Our Approach}
\label{approach}

In this paper, we explore the natural language understanding capabilities of the LLMs by predicting the entity-specific sentiment label from news articles in zero-shot and few-shot settings.

 \subsection{Zero-shot approach} In zero-shot settings, we do not utilize any training exemplar for model supervision. In the absence of demonstration exemplars, the LLM is provided with the prompt containing the problem definition, input context,  and sentiment class labels. Problem definition denotes the task name (i.e. sentiment classification); Input context contains the news text and the target entity name; The sentiment class labels contain the set of final sentiment tags. The prompt defines the expected structure of the output that eventually helps us to decode the LLM-responses into our desired format.
 In our experiment, we utilize two prompting techniques for zero-shot sentiment classification. 

\begin{figure*}[htbp]
  \centering
  \includegraphics[width=0.8\textwidth]{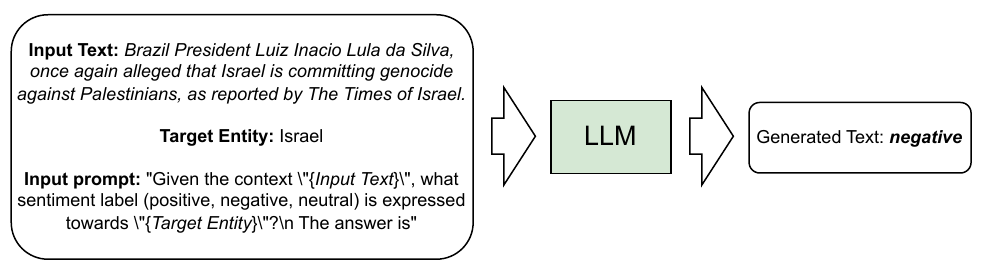}
  \caption{Standard Zero-shot Prompting Demonstration. Input prompt includes news article excerpt and target entity phrase, generating entity-specific sentiment output.}
  \label{fig:std_zero}
\end{figure*}

 \paragraph{standard zero-shot} In standard zero-shot setting, the input prompt contains the task definition, query text and target entity as described in the figure \ref{fig:std_zero}. Due to clarity and space constraint, we present a single sentence instead of the whole document as the input text in that figure.

\begin{figure*}[htbp]
  \centering
  \includegraphics[width=0.95\textwidth]{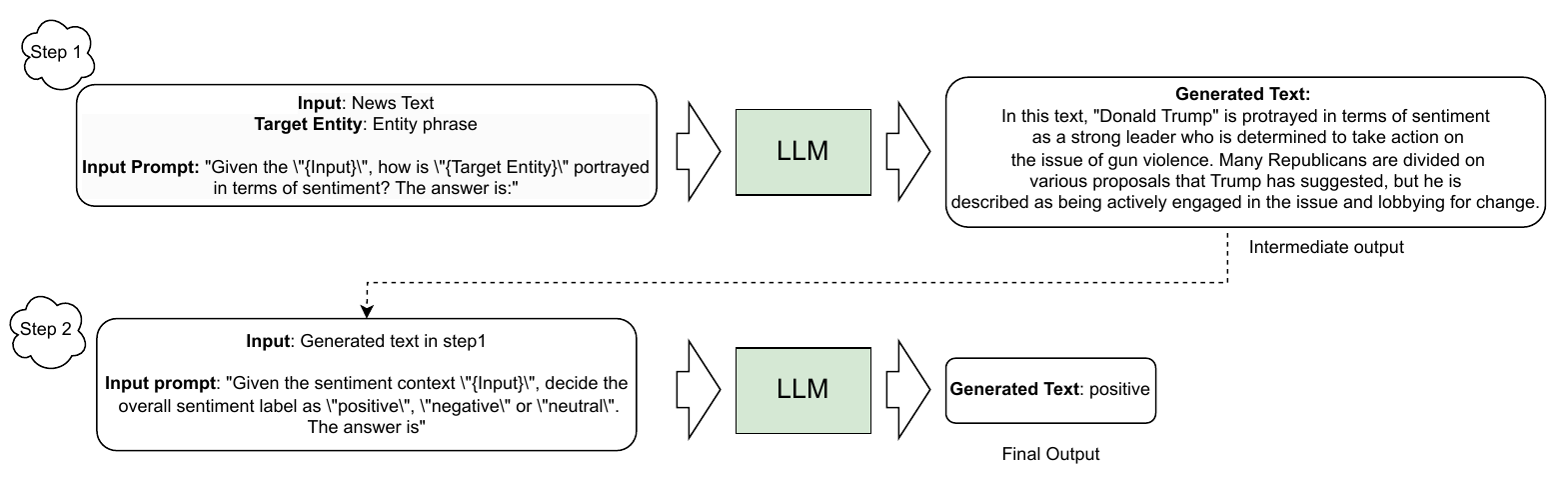}
  \caption{Two-Stage Prompting for Sentiment Prediction in Zero-Shot Setting. The first stage involves extracting rationales for entity-specific sentiment prediction, providing opinions regarding the target entity. In the second stage, sentiment tags are predicted based on the explanations}
  \label{fig:zero-2stage}
\end{figure*}

\paragraph{2-stage Prompting} To comprehend the sentiment towards a particular entity, it is essential to discern both implicit and explicit opinions within the news context concerning that entity. Recent studies, such as \cite{Kojima2022LargeLM}, have demonstrated that a two-stage prompting strategy, referred to as zero-shot Chain of Thought (COT), can enhance the performance of Large Language Models (LLMs) in various reasoning tasks. Consequently, we adopt a similar two-stage prompting approach in our methodology. In the first stage, we extract textual cues indicating how the target entity is depicted sentiment-wise. Subsequently, in the second stage, we predict the final sentiment label. By employing this dual prompting process, we obtain the sentiment label pertaining to the target entity. Figure \ref{fig:zero-2stage} illustrates the 2-stage prompting method, depicting both the intermediate and final outputs. The intermediate output serves as an explanation for the final sentiment prediction.

\subsection{Few-shot approach} \cite{brown2020language} have shown that LLMs can perform new tasks during inference when prompted with a few-demonstrations. In our experiment, we follow different prompting strategies and measure the LLM performance on few-shot scenario.

\begin{figure*}[htbp]
  \centering
  \includegraphics[width=0.95\textwidth]{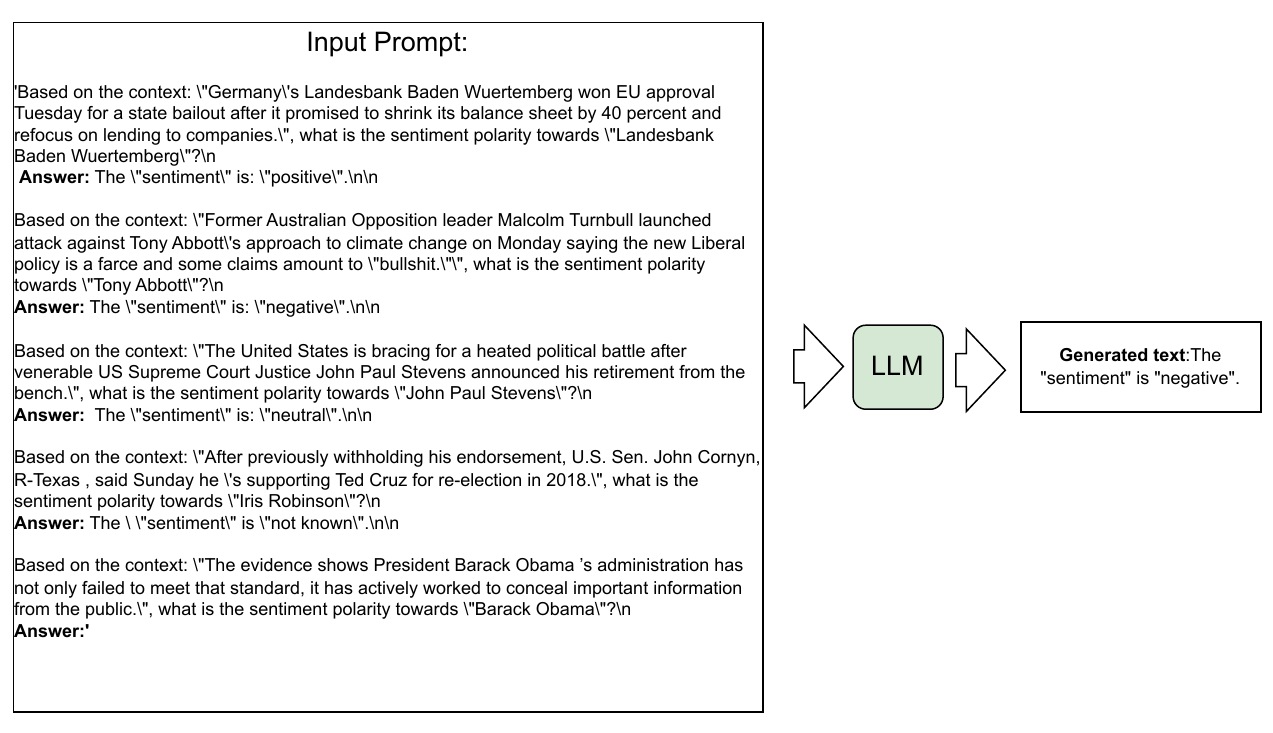}
  \caption{Standard Few-shot Prompting Illustration. Input: Entity context (news article). Output: Entity-specific sentiment tag (positive/negative/neutral).}
  \label{fig:std_few-shot}
\end{figure*}

\paragraph{Standard few-shot} In standard few-shot prompting, the LLMs are provided with in-context exemplars containing (input, output) pairs before providing the query input text. Here, the input is the entity context (news article) and the output is the entity-specific sentiment tag (positive, negative or neutral). The sample input prompt as well as the generated output is reported in the figure \ref{fig:std_few-shot}.

\begin{figure*}[htbp]
  \centering
  \includegraphics[width=0.95\textwidth]{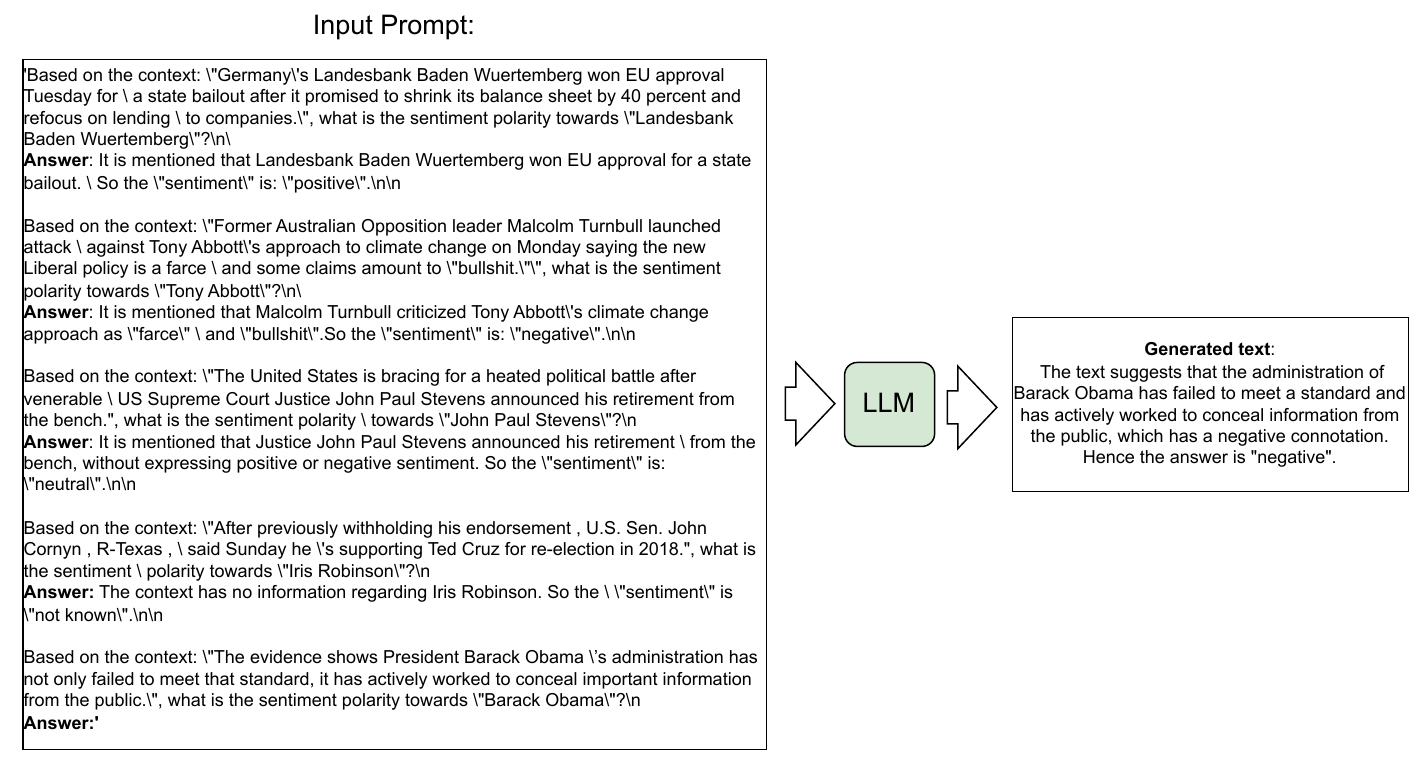}
  \caption{COT Prompting Demonstration. Input: Entity context (news article), Rationale: Justification of sentiment tag, Output: Entity-specific sentiment (positive/negative/neutral).}
  \label{fig:cot_few-shot}
\end{figure*}

\paragraph{COT prompting} In COT prompting, we augment the (input, output) example-pair demonstrations with a natural language rationale that demonstrates the justification of the output sentiment tag. Hence the prompt is a triplet containing (input, rationale, output). By incorporating rationale, we aim is to investigate whether LLMs can benefit from these explanations when learning from exemplars in-context. The sample input prompt is illustrated in figure \ref{fig:cot_few-shot}.

\subsection{Self-Consistency} 
\begin{figure*}[htbp]
  \centering
  \includegraphics[width=0.95\textwidth]{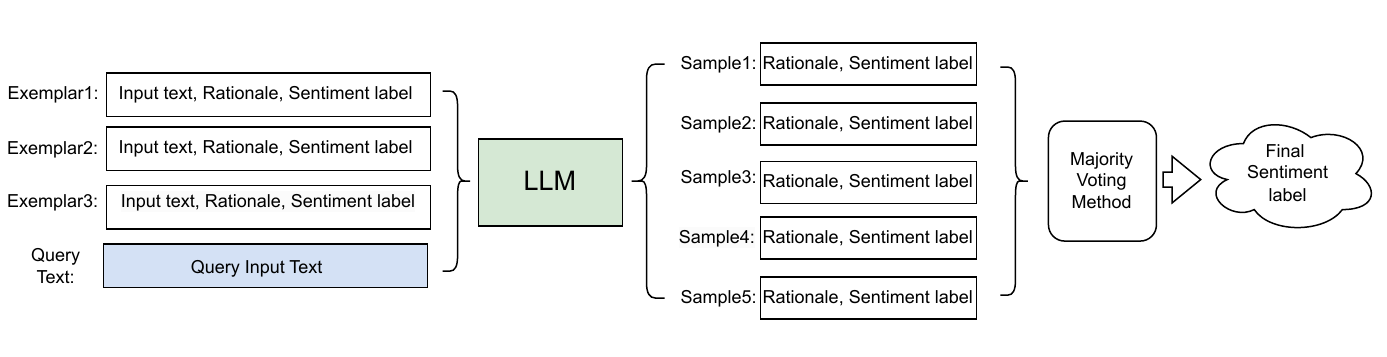}
  \caption{Chain-of-Thought Prompting with Self-Consistency in few-shot settings. The image illustrates the process where the LLM is provided with demonstration exemplars consisting of (input, output, and rationale) triplets. During inference, given a query input, the LLM returns multiple outputs containing sentiment tags and related explanations. The final sentiment tag is identified through sampling and marginalization technique.}
  \label{fig:selfC}
\end{figure*}

Upon scrutinizing news articles, it becomes evident that the portrayal of the same entity varies across different paragraphs within a single article. Hence, to ascertain the overall document-level sentiment of the target entity, it is imperative to encompass and weigh all these opinions pertaining to the entity. \cite{Wang2022SelfConsistencyIC}, introduced the concept of self-consistency, which revolves around generating multiple reasoning paths to determine the correct final answer. Leveraging this concept, we aim to influence the decoder of Large Language Models (LLMs) to produce a diverse set of reasoning paths for predicting the final sentiment tag. Following \cite{Wang2022SelfConsistencyIC}, in the self-consistency method, we first prompt the LLM using chain-of-thought prompting. Subsequently, instead of employing a greedy search decoding approach, we utilize various existing sampling algorithms to generate a diverse set of candidate reasoning paths. Each of these paths may lead to a different final sentiment label. Finally, employing a majority voting approach, we marginalize the sampled reasoning paths and select the sentiment label that remains consistent across all generated answers. Through this "sample-and-marginalize" decoding method, our objective is to encapsulate all sentiment-inducing components from the news content and amalgamate them to determine the overall sentiment. The self-consistency method is depicted in Figure \ref{fig:selfC}.

\section{Experimental Setup}
\subsection{Datasets} In our experiment, we utilize datasets released by \cite{bastan2020author} for our model evaluation. Additionally, we curate a news dataset focused on the political domain, extracted and annotated by the Event-Registry API. These datasets consist of news articles, target entity phrases, and entity-specific sentiment tags, but lack rationale information. In the following subsections, we provide detailed descriptions of these two datasets.

\paragraph{PerSenT:} The PerSenT dataset, introduced by \cite{bastan2020author}, is designed to predict the author's sentiment towards the main entities in news articles. This dataset includes paragraph-level as well as entire document-level sentiment annotations towards the target entity. The authors have partitioned the entire dataset into train, development, and test splits. Additionally, the paper reports the performance of various fine-tuned BERT model variants on this dataset. In our experiment, we evaluate the performance of our proposed LLM models on the test set of the PerSenT dataset.

\begin{table}[!htp]
\center
\fontsize{7}{11}\selectfont
\setlength\tabcolsep{5pt}
\begin{tabular}{c|c|c|c|c|c}
\hline
Dataset                                                     & Positive & Neutral & Negative & Total & \begin{tabular}[c]{@{}c@{}}Unique\\ Entites\end{tabular} \\ \hline
\begin{tabular}[c]{@{}c@{}}PerSenT\\ Test-Std\end{tabular}  & 293      & 213     & 73       & 579   & 426                                                      \\ \hline
\begin{tabular}[c]{@{}c@{}}PerSenT\\ Test-Freq\end{tabular} & 368      & 320     & 139      & 827   & 4                                                        \\ \hline
WPAN                                                        & 600      & 600     & 600      & 1800  & 3                                                        \\ \hline
\end{tabular}

\caption{Comparison of Test Dataset Statistics: PerSenT vs. WPAN for Entity-Specific Sentiment Analysis}
\label{table:datastat}
\end{table}

\paragraph{Event Registry Data:} The PerSenT dataset mentioned earlier contains a diverse selection of news articles from various domains. However, to explore how different countries and their policies are portrayed in the media, we have compiled a new dataset focused on global politics. This involved gathering news articles pertaining to specific nations and their policies from media outlets worldwide. We named the dataset as WPAN (\textbf{W}orldwide \textbf{P}erception \textbf{A}nalysis of \textbf{N}ations)\footnote{\url{https://github.com/alapanju/EntSent}}. The selected nations include India, Russia, and Israel. For India, we included media sources from neighboring countries such as Pakistan and Bangladesh. Similarly, for Russia, we selected outlets from India, the UK, and the USA, while for Israel, we included outlets from India and Pakistan. Our selection criteria were based on the significance of these nations in events such as the Russia-Ukraine conflict, the Israel-Hamas conflict, and tensions in the Indian subcontinent. We chose media sources based on article frequency, ensuring a random selection without bias. We used the Event-Registry Python API to extract relevant news articles. The API also provides a target specific document level sentiment score. The sentiment scores range from $-1$ to $1$, where $1$ represents maximum positive sentiment. We categorized sentiment scores between $0.6$ to $1$ as positive, $-0.2$ to $+0.2$ as neutral, and -0.6 to -1 as negative. We collected 200 articles for each sentiment range for each target topic, resulting in $600$ articles per topic. Each dataset record includes the news article, target entity, and entity-specific sentiment label.

\subsection{Language Models}

Our task integrates entity-specific sentiment analysis with elements of reasoning, particularly in justifying assigned sentiment based on contextual cues and linguistic patterns in the text. Traditional deep learning models or small language models (SLMs) require rationale-augmented training, which is costly and challenging to scale. However, in-context few-shot learning via prompting and the COT approach significantly enhance LLMs' reasoning capabilities across various tasks. Through COT, LLMs can perform few-shot prompting for reasoning tasks using triplets: (input, output, chain-of-thought). Studies have demonstrated the effectiveness of COT in improving reasoning abilities in LLMs with large parameter sizes, which are often inaccessible due to resource constraints. In this study, we aim to determine whether LLMs with fewer parameters can generate explicit reasoning while predicting entity-specific sentiment classes from document context. To explore this, we experiment with three transformer-based language models of varying scales:

\begin{itemize}

    \item Mistral \cite{jiang2023mistral} is a open-sourced decoder-based model with 7-billion parameters.
    
    \item Llama-2 \cite{touvron2023llama}, developed by Meta AI, is a collection of transformer-based language models ranging in scale from 7 billion to 70 billion parameters. In our experiment, we use fine-tuned model named Llama2-13b-chat-hf with 13-billion parameters.
    
    \item Falcon \cite{Almazrouei2023TheFS} is a causal decoder-only open-sourced language model. In our experiment, we use \textit{instruct} version of the language model with two different parameter size namely, Falcon-7b-instruct (7-billion) and Falcon-40b-instruct (40-billion).

\end{itemize}

\subsection{Prompting and Decoding Scheme} We discuss about the prompt designing in the section \ref{approach}. For few-shot setting, we employ 3-4 exemplars in our experiment. These samples are chosen randomly from the training set of the PerSenT dataset. In COT prompting method, the exemplar pairs are augmented with manually composed natural language explanations. For a fair comparison, we use same prompt structure and same exemplar sets across all the LLMs.

For standard zero-shot and few-shot methods, we use greedy search decoding. In the case of 2-stage prompting method, we first use top-k, top-p and temperature sampling and in the second step, we employ greedy decoding method. In self-consistency method, instead of greedy search decoding, we employ different sampling algorithms like top-$k$ sampling, temperature sampling, top-$p$ sampling. In our experiment, the range of $k$ value is between$\{50, 90\}$; The $p$ value varies in the range of $\{0.9,0.95\}$. We keep the temperature value as $0.7$. 
We utilized a BERT model~\cite{Devlin2019BERTPO} fine-tuned with the training data from the PerSent dataset as our baseline.

\begin{table*}[!htp]
\begin{tabular}{|c|cccccc|}
\hline
                    & \multicolumn{3}{c|}{Zero-shot}                                                                                                                                                                                                                    & \multicolumn{3}{c|}{Few-shot}                                                                                                                                                                                            \\ \hline
Model               & \multicolumn{1}{c|}{\begin{tabular}[c]{@{}c@{}}Std\\ decoding\end{tabular}} & \multicolumn{1}{c|}{\begin{tabular}[c]{@{}c@{}}2-stage\\ prompting\end{tabular}} & \multicolumn{1}{c|}{\begin{tabular}[c]{@{}c@{}}Self-\\ Consistency\end{tabular}} & \multicolumn{1}{c|}{\begin{tabular}[c]{@{}c@{}}Std\\ few-shot\end{tabular}} & \multicolumn{1}{c|}{\begin{tabular}[c]{@{}c@{}}COT\\ prompting\end{tabular}} & \begin{tabular}[c]{@{}c@{}}Self-\\ Consistency\end{tabular} \\ \hline
Mistral-7b          & \multicolumn{1}{c|}{42.16}                                                  & \multicolumn{1}{c|}{43.32}                                                       & \multicolumn{1}{c|}{45.67}                                                       & \multicolumn{1}{c|}{49.87}                                                  & \multicolumn{1}{c|}{49.56}                                                   & 52.78                                                       \\ \hline
Llama-13b-chat      & \multicolumn{1}{c|}{41.59}                                                  & \multicolumn{1}{c|}{42.92}                                                       & \multicolumn{1}{c|}{43.21}                                                       & \multicolumn{1}{c|}{49.43}                                                  & \multicolumn{1}{c|}{50.88}                                                   & 51.97                                                       \\ \hline
Falcon-7b-instruct  & \multicolumn{1}{c|}{41.47}                                                  & \multicolumn{1}{c|}{42.81}                                                       & \multicolumn{1}{c|}{44.22}                                                       & \multicolumn{1}{c|}{48.56}                                                  & \multicolumn{1}{c|}{49.87}                                                   & 52.63                                                       \\ \hline
Falcon-40b-instruct & \multicolumn{1}{c|}{43.89}                                                  & \multicolumn{1}{c|}{44.13}                                                       & \multicolumn{1}{c|}{47.05}                                                       & \multicolumn{1}{c|}{50.24}                                                  & \multicolumn{1}{c|}{51.39}                                                   & 54.94                                                       \\ \hline
Fine-tuned BERT     & \multicolumn{6}{c|}{43.07}                                                                                                                                                                                                                                                                                                                                                                                                                                                   \\ \hline
\end{tabular}
\caption{Macro F1-score for Document-level Entity-centric Sentiment prediction across Various LLMs on the PerSenT \textit{Frequent} Test Dataset}
\label{result_persent_frq}
\end{table*}

\begin{table*}[!htp]
\begin{tabular}{|c|cccccc|}
\hline
                    & \multicolumn{3}{c|}{Zero-shot}                                                                                                                                                                                                                    & \multicolumn{3}{c|}{Few-shot}                                                                                                                                                                                            \\ \hline
Model               & \multicolumn{1}{c|}{\begin{tabular}[c]{@{}c@{}}Std\\ decoding\end{tabular}} & \multicolumn{1}{c|}{\begin{tabular}[c]{@{}c@{}}2-stage\\ prompting\end{tabular}} & \multicolumn{1}{c|}{\begin{tabular}[c]{@{}c@{}}Self-\\ Consistency\end{tabular}} & \multicolumn{1}{c|}{\begin{tabular}[c]{@{}c@{}}Std\\ few-shot\end{tabular}} & \multicolumn{1}{c|}{\begin{tabular}[c]{@{}c@{}}COT\\ prompting\end{tabular}} & \begin{tabular}[c]{@{}c@{}}Self-\\ Consistency\end{tabular} \\ \hline
Mistral-7b          & \multicolumn{1}{c|}{43.61}                                                  & \multicolumn{1}{c|}{44.29}                                                       & \multicolumn{1}{c|}{44.70}                                                       & \multicolumn{1}{c|}{46.64}                                                  & \multicolumn{1}{c|}{47.37}                                                   & 49.77                                                       \\ \hline
Llama-13b-chat      & \multicolumn{1}{c|}{42.09}                                                  & \multicolumn{1}{c|}{43.61}                                                       & \multicolumn{1}{c|}{44.92}                                                       & \multicolumn{1}{c|}{45.98}                                                  & \multicolumn{1}{c|}{46.56}                                                   & 49.08                                                       \\ \hline
Falcon-7b-instruct  & \multicolumn{1}{c|}{41.72}                                                  & \multicolumn{1}{c|}{42.19}                                                       & \multicolumn{1}{c|}{43.64}                                                       & \multicolumn{1}{c|}{47.16}                                                  & \multicolumn{1}{c|}{48.07}                                                   & 50.45                                                       \\ \hline
Falcon-40b-instruct & \multicolumn{1}{c|}{44.21}                                                  & \multicolumn{1}{c|}{45.27}                                                       & \multicolumn{1}{c|}{46.17}                                                       & \multicolumn{1}{c|}{49.67}                                                  & \multicolumn{1}{c|}{51.19}                                                   & 53.71                                                       \\ \hline
Fine-tuned BERT     & \multicolumn{6}{c|}{48.38}                                                                                                                                                                                                                                                                                                                                                                                                                                                   \\ \hline
\end{tabular}
\caption{Macro F1-score for Document-level Entity-centric Sentiment prediction across Various LLMs on the PerSenT \textit{Standard} Test Dataset}
\label{result_persent_std}
\end{table*}

\begin{table*}[!htp]
\begin{tabular}{|c|cccccc|}
\hline
                    & \multicolumn{3}{c|}{Zero-shot}                                                                                                                                                                                                                    & \multicolumn{3}{c|}{Few-shot}                                                                                                                                                                                            \\ \hline
Model               & \multicolumn{1}{c|}{\begin{tabular}[c]{@{}c@{}}Std\\ decoding\end{tabular}} & \multicolumn{1}{c|}{\begin{tabular}[c]{@{}c@{}}2-stage\\ prompting\end{tabular}} & \multicolumn{1}{c|}{\begin{tabular}[c]{@{}c@{}}Self-\\ Consistency\end{tabular}} & \multicolumn{1}{c|}{\begin{tabular}[c]{@{}c@{}}Std\\ few-shot\end{tabular}} & \multicolumn{1}{c|}{\begin{tabular}[c]{@{}c@{}}COT\\ prompting\end{tabular}} & \begin{tabular}[c]{@{}c@{}}Self-\\ Consistency\end{tabular} \\ \hline
Mistral-7b          & \multicolumn{1}{c|}{56.67}                                                  & \multicolumn{1}{c|}{58.12}                                                       & \multicolumn{1}{c|}{59.76}                                                       & \multicolumn{1}{c|}{59.07}                                                  & \multicolumn{1}{c|}{59.36}                                                   & 61.49                                                       \\ \hline
Llama-13b-chat      & \multicolumn{1}{c|}{54.43}                                                  & \multicolumn{1}{c|}{57.07}                                                       & \multicolumn{1}{c|}{58.21}                                                       & \multicolumn{1}{c|}{58.34}                                                  & \multicolumn{1}{c|}{58.42}                                                   & 59.79                                                       \\ \hline
Falcon-7b-instruct  & \multicolumn{1}{c|}{55.71}                                                  & \multicolumn{1}{c|}{57.18}                                                       & \multicolumn{1}{c|}{58.23}                                                       & \multicolumn{1}{c|}{59.69}                                                  & \multicolumn{1}{c|}{59.35}                                                   & 61.48                                                       \\ \hline
Falcon-40b-instruct & \multicolumn{1}{c|}{57.95}                                                  & \multicolumn{1}{c|}{59.01}                                                       & \multicolumn{1}{c|}{59.91}                                                       & \multicolumn{1}{c|}{61.84}                                                  & \multicolumn{1}{c|}{62.07}                                                   & 63.87                                                       \\ \hline
Fine-tuned BERT     & \multicolumn{6}{c|}{56.54}                                                                                                                                                                                                                                                                                                                                                                                                                                                   \\ \hline
\end{tabular}
\caption{Macro F1-score for Document-level Entity-centric Sentiment prediction across Various LLMs on the WPAN Dataset}
\label{result_wpan}
\end{table*}

\section{Result Analysis}

In this section, we evaluate the performance of the Large Language Models (LLMs) based on the correctness of the final predicted sentiment labels using macro-F1 score metric for quantitative analysis. The experiments are conducted multiple times with different sets of training samples, and the average output over three runs is reported to ensure consistency. The seed value is fixed during experiments to obtain identical outputs.

Our analysis reveals variations in LLM performance between datasets. Specifically, the LLMs perform better on the WPAN dataset compared to the PerSenT dataset. Upon examining the news articles, we observed that most documents in the PerSenT dataset exhibit mixed sentiment across various paragraphs within the same article. In contrast, the sentiment across paragraphs in the WPAN dataset is less varied, potentially contributing to the improved performance of LLMs on this dataset.

We address three main experimental questions during the result analysis:

\begin{itemize}
    \item We investigate whether LLMs can predict entity-specific sentiment labels in a zero-shot setting.
    \item We explore whether LLMs can learn from few-shot demonstrations.
    \item We analyze whether scaling up the LLM size has any effect on zero-shot and few-shot settings.

\end{itemize}

The experimental results, presented in Tables \ref{result_persent_frq}, \ref{result_persent_std}, and \ref{result_wpan}, reveal several key insights.

Firstly, for the PerSenT-freq and WPAN datasets, the FALCON-40b model consistently outperforms the fine-tuned BERT model. Even in the case of PerSenT-std data, the Self-consistency method over FALCON-40b yields comparable performance compared to FT-BERT. These findings indicate that LLMs, with their pretraining and proper parameter sizing, exhibit a strong capability to capture sentiment labels from documents in zero-shot settings.

Secondly, we observe that the model performance improves significantly in few-shot settings compared to zero-shot scenarios across all three datasets. This suggests that learning in-context positively impacts model performance and effectiveness.

However, our experiments also reveal that the Chain-of-Thought (COT) prompting method is not consistently effective. In some cases, its performance lags behind standard few-shot approaches. Nevertheless, the self-consistency method proves to be beneficial in enhancing model performance across all cases. Additionally, in zero-shot scenarios, the 2-stage prompting approach outperforms the standard zero-shot method.

Lastly, we experimented with LLMs having a parameter size within 40-billion. However, at this scale, we did not observe significant effects of model scaling on performance. Further exploration with larger model sizes may provide additional insights into this aspect.

Overall, our results demonstrate the effectiveness of LLMs in entity-specific sentiment prediction, particularly in few-shot learning scenarios, while highlighting the importance of appropriate prompting strategies and model architectures.

\section{Related Work}

\subsection{Sentiment Analysis in news domain} Sentiment analysis is a widely explored area within natural language processing, attracting significant attention due to its multitude of applications across academic research and practical domains. Particularly within the realm of news content analysis, sentiment analysis has emerged as a pivotal task~\cite{balahur2013sentiment, KATAYAMA20191287, Islam2017PolarityDO, kuila2024analyzing, 10.1145/3230348.3230354, Pryzant2019AutomaticallyNS}. Researchers have also delved into sentiment prediction grounded in news events~\cite{zhou-etal-2021-implicit}. Moreover, there is a burgeoning interest in news bias analysis~\cite{doi:10.1177/0093650215614364}, which often relies on sentiment associated with news publications~\cite{RODRIGOGINES2024121641}.

However, our focus in this paper lies specifically on entity-specific sentiment analysis within news articles. This presents a distinct problem with diverse applications, including predicting authors' sentiment~\cite{bastan2020author}, discerning the ideology of news outlets~\cite{Lin2011MoreVT}, and analyzing media bias~\cite{8791197}.

\subsection{Large Language Models} 
Recent advancements in natural language processing (NLP) research have been marked by the emergence of Large Language Models (LLMs)~\cite{Chowdhery2022PaLMSL}. These LLMs are pre-trained on massive text corpora using diverse training techniques such as instruction-tuning and reinforcement learning with human feedback (RLHF)~\cite{Christiano2017DeepRL}, showcasing impressive performance in zero-shot and few-shot settings. The paradigm shift towards in-context learning~\cite{brown2020language} has further enhanced the capabilities of LLMs, moving away from fine-tuning to prompt-tuning approaches.

Despite the widespread adoption of LLMs in various NLP tasks, including sentiment analysis~\cite{Zhong2023CanCU, Wang2023IsCA}, challenges persist due to the computational demands of these models, particularly in resource-constrained settings. Consequently, our research is dedicated to investigating the viability of employing smaller-scale LLMs for entity-specific sentiment identification within the domain of political news articles. By harnessing the power of these compact LLMs, we endeavor to address resource limitations while leveraging the inherent capabilities of LLMs in sentiment analysis tasks.

\section{Conclusion}

In this study, we investigated the application of Large Language Models (LLMs) in predicting entity-specific sentiment from political news articles using zero-shot and few-shot strategies. Our findings demonstrate the effectiveness of LLMs, particularly FALCON-40b, in capturing sentiment towards political entities. Leveraging the chain-of-thought (COT) approach with rationale in few-shot in-context learning, we observed improvements in sentiment prediction accuracy, especially in few-shot scenarios. While the self-consistency mechanism enhanced consistency in sentiment prediction, we noted varying effectiveness in the COT prompting method across different datasets. Overall, our results highlight the potential of LLMs in entity-centric sentiment analysis within the political news domain.

Beyond sentiment analysis, our work has broader implications for media bias analysis and identification of media house ideologies. By discerning sentiment towards political entities, our model can assist in analyzing media bias and understanding the ideological stance of media houses. This capability holds promise for enhancing media literacy and facilitating informed discourse in political communication.

Moving forward, future research could explore additional applications of LLMs in political NLP tasks, such as misinformation detection, stance classification, and agenda setting analysis. Additionally, investigating the interpretability of LLMs' predictions and addressing potential biases in training data are essential considerations for further advancement in this field.

In conclusion, our study contributes to advancing the understanding of sentiment analysis in the political news domain and underscores the potential of LLMs in facilitating nuanced analysis of media content and political discourse.


 \section{Bibliographical References}\label{sec:reference}

\bibliographystyle{lrec-coling2024-natbib}
\bibliography{lrec-coling2024-example}

\begin{thebibliography}{0}
\expandafter\ifx\csname natexlab\endcsname\relax\def\natexlab#1{#1}\fi

\end{thebibliography}


\begin{thebibliography}{40}
\expandafter\ifx\csname natexlab\endcsname\relax\def\natexlab#1{#1}\fi

\bibitem[{Almazrouei et~al.(2023)Almazrouei, Alobeidli, Alshamsi, Cappelli,
  Cojocaru, Hesslow, Launay, Malartic, Mazzotta, Noune, Pannier, and
  Penedo}]{Almazrouei2023TheFS}
Ebtesam Almazrouei, Hamza Alobeidli, Abdulaziz Alshamsi, Alessandro Cappelli,
  Ruxandra-Aim{\'e}e Cojocaru, Daniel Hesslow, Julien Launay, Quentin Malartic,
  Daniele Mazzotta, Badreddine Noune, Baptiste Pannier, and Guilherme Penedo.
  2023.
\newblock \href {https://api.semanticscholar.org/CorpusID:265466629} {The
  falcon series of open language models}.
\newblock \emph{ArXiv}, abs/2311.16867.

\bibitem[{Balahur et~al.(2013)Balahur, Steinberger, Kabadjov, Zavarella, Van
  Der~Goot, Halkia, Pouliquen, and Belyaeva}]{balahur2013sentiment}
Alexandra Balahur, Ralf Steinberger, Mijail Kabadjov, Vanni Zavarella, Erik Van
  Der~Goot, Matina Halkia, Bruno Pouliquen, and Jenya Belyaeva. 2013.
\newblock Sentiment analysis in the news.
\newblock \emph{arXiv preprint arXiv:1309.6202}.

\bibitem[{Bastan et~al.(2020)Bastan, Koupaee, Son, Sicoli, and
  Balasubramanian}]{bastan2020author}
Mohaddeseh Bastan, Mahnaz Koupaee, Youngseo Son, Richard Sicoli, and Niranjan
  Balasubramanian. 2020.
\newblock Author's sentiment prediction.
\newblock \emph{arXiv preprint arXiv:2011.06128}.

\bibitem[{Brown et~al.(2020)Brown, Mann, Ryder, Subbiah, Kaplan, Dhariwal,
  Neelakantan, Shyam, Sastry, Askell et~al.}]{brown2020language}
Tom Brown, Benjamin Mann, Nick Ryder, Melanie Subbiah, Jared~D Kaplan, Prafulla
  Dhariwal, Arvind Neelakantan, Pranav Shyam, Girish Sastry, Amanda Askell,
  et~al. 2020.
\newblock Language models are few-shot learners.
\newblock \emph{Advances in neural information processing systems},
  33:1877--1901.

\bibitem[{Brun and Nikoulina(2018)}]{brun2018aspect}
Caroline Brun and Vassilina Nikoulina. 2018.
\newblock Aspect based sentiment analysis into the wild.
\newblock In \emph{Proceedings of the 9th workshop on computational approaches
  to subjectivity, sentiment and social media analysis}, pages 116--122.

\bibitem[{Chowdhery et~al.(2022)Chowdhery, Narang, Devlin, Bosma, Mishra,
  Roberts, Barham, Chung, Sutton, Gehrmann, Schuh, Shi, Tsvyashchenko, Maynez,
  Rao, Barnes, Tay, Shazeer, Prabhakaran, Reif, Du, Hutchinson, Pope, Bradbury,
  Austin, Isard, Gur-Ari, Yin, Duke, Levskaya, Ghemawat, Dev, Michalewski,
  Garc{\'i}a, Misra, Robinson, Fedus, Zhou, Ippolito, Luan, Lim, Zoph,
  Spiridonov, Sepassi, Dohan, Agrawal, Omernick, Dai, Pillai, Pellat,
  Lewkowycz, Moreira, Child, Polozov, Lee, Zhou, Wang, Saeta, Diaz, Firat,
  Catasta, Wei, Meier-Hellstern, Eck, Dean, Petrov, and
  Fiedel}]{Chowdhery2022PaLMSL}
Aakanksha Chowdhery, Sharan Narang, Jacob Devlin, Maarten Bosma, Gaurav Mishra,
  Adam Roberts, Paul Barham, Hyung~Won Chung, Charles Sutton, Sebastian
  Gehrmann, Parker Schuh, Kensen Shi, Sasha Tsvyashchenko, Joshua Maynez,
  Abhishek Rao, Parker Barnes, Yi~Tay, Noam~M. Shazeer, Vinodkumar Prabhakaran,
  Emily Reif, Nan Du, Benton~C. Hutchinson, Reiner Pope, James Bradbury, Jacob
  Austin, Michael Isard, Guy Gur-Ari, Pengcheng Yin, Toju Duke, Anselm
  Levskaya, Sanjay Ghemawat, Sunipa Dev, Henryk Michalewski, Xavier Garc{\'i}a,
  Vedant Misra, Kevin Robinson, Liam Fedus, Denny Zhou, Daphne Ippolito, David
  Luan, Hyeontaek Lim, Barret Zoph, Alexander Spiridonov, Ryan Sepassi, David
  Dohan, Shivani Agrawal, Mark Omernick, Andrew~M. Dai,
  Thanumalayan~Sankaranarayana Pillai, Marie Pellat, Aitor Lewkowycz,
  Erica~Oliveira Moreira, Rewon Child, Oleksandr Polozov, Katherine Lee,
  Zongwei Zhou, Xuezhi Wang, Brennan Saeta, Mark Diaz, Orhan Firat, Michele
  Catasta, Jason Wei, Kathleen~S. Meier-Hellstern, Douglas Eck, Jeff Dean, Slav
  Petrov, and Noah Fiedel. 2022.
\newblock Palm: Scaling language modeling with pathways.

\bibitem[{Christiano et~al.(2017)Christiano, Leike, Brown, Martic, Legg, and
  Amodei}]{Christiano2017DeepRL}
Paul~Francis Christiano, Jan Leike, Tom~B. Brown, Miljan Martic, Shane Legg,
  and Dario Amodei. 2017.
\newblock \href {https://api.semanticscholar.org/CorpusID:4787508} {Deep
  reinforcement learning from human preferences}.
\newblock \emph{ArXiv}, abs/1706.03741.

\bibitem[{Devlin et~al.(2019)Devlin, Chang, Lee, and
  Toutanova}]{Devlin2019BERTPO}
Jacob Devlin, Ming-Wei Chang, Kenton Lee, and Kristina Toutanova. 2019.
\newblock \href {https://api.semanticscholar.org/CorpusID:52967399} {Bert:
  Pre-training of deep bidirectional transformers for language understanding}.
\newblock In \emph{North American Chapter of the Association for Computational
  Linguistics}.

\bibitem[{Eberl et~al.(2017)Eberl, Boomgaarden, and
  Wagner}]{doi:10.1177/0093650215614364}
Jakob-Moritz Eberl, Hajo~G. Boomgaarden, and Markus Wagner. 2017.
\newblock \href {https://doi.org/10.1177/0093650215614364} {One bias fits all?
  three types of media bias and their effects on party preferences}.
\newblock \emph{Communication Research}, 44(8):1125--1148.

\bibitem[{Fei et~al.(2023)Fei, Li, Liu, Bing, Li, and seng
  Chua}]{Fei2023ReasoningIS}
Hao Fei, Bobo Li, Qian Liu, Lidong Bing, Fei Li, and Tat seng Chua. 2023.
\newblock \href {https://api.semanticscholar.org/CorpusID:258822870} {Reasoning
  implicit sentiment with chain-of-thought prompting}.
\newblock In \emph{Annual Meeting of the Association for Computational
  Linguistics}.

\bibitem[{Golovanov et~al.(2019)Golovanov, Kurbanov, Nikolenko, Truskovskyi,
  Tselousov, and Wolf}]{golovanov-etal-2019-large}
Sergey Golovanov, Rauf Kurbanov, Sergey Nikolenko, Kyryl Truskovskyi, Alexander
  Tselousov, and Thomas Wolf. 2019.
\newblock \href {https://doi.org/10.18653/v1/P19-1608} {Large-scale transfer
  learning for natural language generation}.
\newblock In \emph{Proceedings of the 57th Annual Meeting of the Association
  for Computational Linguistics}, pages 6053--6058, Florence, Italy.
  Association for Computational Linguistics.

\bibitem[{Hamborg et~al.(2019)Hamborg, Zhukova, and Gipp}]{8791197}
Felix Hamborg, Anastasia Zhukova, and Bela Gipp. 2019.
\newblock \href {https://doi.org/10.1109/JCDL.2019.00036} {Automated
  identification of media bias by word choice and labeling in news articles}.
\newblock In \emph{2019 ACM/IEEE Joint Conference on Digital Libraries (JCDL)},
  pages 196--205.

\bibitem[{Huang and Chang(2022)}]{huang2022towards}
Jie Huang and Kevin Chen-Chuan Chang. 2022.
\newblock Towards reasoning in large language models: A survey.
\newblock \emph{arXiv preprint arXiv:2212.10403}.

\bibitem[{Islam et~al.(2017)Islam, Ashraf, Abir, and
  Mottalib}]{Islam2017PolarityDO}
Muhammad~Usama Islam, Faisal~Bin Ashraf, Ali~Imam Abir, and M.~Abdul Mottalib.
  2017.
\newblock \href {https://api.semanticscholar.org/CorpusID:22751214} {Polarity
  detection of online news articles based on sentence structure and dynamic
  dictionary}.
\newblock \emph{2017 20th International Conference of Computer and Information
  Technology (ICCIT)}, pages 1--5.

\bibitem[{Jiang et~al.(2023)Jiang, Sablayrolles, Mensch, Bamford, Chaplot,
  de~las Casas, Bressand, Lengyel, Lample, Saulnier, Lavaud, Lachaux, Stock,
  Scao, Lavril, Wang, Lacroix, and Sayed}]{jiang2023mistral}
Albert~Q. Jiang, Alexandre Sablayrolles, Arthur Mensch, Chris Bamford,
  Devendra~Singh Chaplot, Diego de~las Casas, Florian Bressand, Gianna Lengyel,
  Guillaume Lample, Lucile Saulnier, Lélio~Renard Lavaud, Marie-Anne Lachaux,
  Pierre Stock, Teven~Le Scao, Thibaut Lavril, Thomas Wang, Timothée Lacroix,
  and William~El Sayed. 2023.
\newblock \href {http://arxiv.org/abs/2310.06825} {Mistral 7b}.

\bibitem[{Katayama et~al.(2019)Katayama, Kino, and Tsuda}]{KATAYAMA20191287}
Daisuke Katayama, Yasunobu Kino, and Kazuhiko Tsuda. 2019.
\newblock \href {https://doi.org/https://doi.org/10.1016/j.procs.2019.09.298}
  {A method of sentiment polarity identification in financial news using deep
  learning}.
\newblock \emph{Procedia Computer Science}, 159:1287--1294.
\newblock Knowledge-Based and Intelligent Information \& Engineering Systems:
  Proceedings of the 23rd International Conference KES2019.

\bibitem[{Kenyon-Dean et~al.(2018)Kenyon-Dean, Ahmed, Fujimoto,
  Georges-Filteau, Glasz, Kaur, Lalande, Bhanderi, Belfer, Kanagasabai,
  Sarrazingendron, Verma, and Ruths}]{kenyon-dean-etal-2018-sentiment}
Kian Kenyon-Dean, Eisha Ahmed, Scott Fujimoto, Jeremy Georges-Filteau,
  Christopher Glasz, Barleen Kaur, Auguste Lalande, Shruti Bhanderi, Robert
  Belfer, Nirmal Kanagasabai, Roman Sarrazingendron, Rohit Verma, and Derek
  Ruths. 2018.
\newblock \href {https://doi.org/10.18653/v1/N18-1171} {Sentiment analysis:
  It{'}s complicated!}
\newblock In \emph{Proceedings of the 2018 Conference of the North {A}merican
  Chapter of the Association for Computational Linguistics: Human Language
  Technologies, Volume 1 (Long Papers)}, pages 1886--1895, New Orleans,
  Louisiana. Association for Computational Linguistics.

\bibitem[{Kojima et~al.(2022)Kojima, Gu, Reid, Matsuo, and
  Iwasawa}]{Kojima2022LargeLM}
Takeshi Kojima, Shixiang~Shane Gu, Machel Reid, Yutaka Matsuo, and Yusuke
  Iwasawa. 2022.
\newblock \href {https://api.semanticscholar.org/CorpusID:249017743} {Large
  language models are zero-shot reasoners}.
\newblock \emph{ArXiv}, abs/2205.11916.

\bibitem[{Kuila et~al.(2024)Kuila, Jena, Sarkar, and
  Chakrabarti}]{kuila2024analyzing}
Alapan Kuila, Somnath Jena, Sudeshna Sarkar, and Partha~Pratim Chakrabarti.
  2024.
\newblock Analyzing sentiment polarity reduction in news presentation through
  contextual perturbation and large language models.
\newblock \emph{arXiv preprint arXiv:2402.02145}.

\bibitem[{Kumaresan and Thangaraju(2023)}]{KUMARESAN2023100663}
C.~Kumaresan and P.~Thangaraju. 2023.
\newblock \href {https://doi.org/https://doi.org/10.1016/j.measen.2022.100663}
  {Elsa: Ensemble learning based sentiment analysis for diversified text}.
\newblock \emph{Measurement: Sensors}, 25:100663.

\bibitem[{Lin et~al.(2011)Lin, Bagrow, and Lazer}]{Lin2011MoreVT}
Y.~Lin, James~P. Bagrow, and David M.~J. Lazer. 2011.
\newblock \href {https://api.semanticscholar.org/CorpusID:16425016} {More
  voices than ever? quantifying media bias in networks}.
\newblock \emph{ArXiv}, abs/1111.1227.

\bibitem[{Liu(2020)}]{liu2020sentiment}
Bing Liu. 2020.
\newblock \emph{Sentiment analysis: Mining opinions, sentiments, and emotions}.
\newblock Cambridge university press.

\bibitem[{Poria et~al.(2020)Poria, Hazarika, Majumder, and
  Mihalcea}]{Poria2020BeneathTT}
Soujanya Poria, Devamanyu Hazarika, Navonil Majumder, and Rada Mihalcea. 2020.
\newblock \href {https://api.semanticscholar.org/CorpusID:218470466} {Beneath
  the tip of the iceberg: Current challenges and new directions in sentiment
  analysis research}.
\newblock \emph{IEEE Transactions on Affective Computing}, 14:108--132.

\bibitem[{Pryzant et~al.(2019)Pryzant, Martinez, Dass, Kurohashi, Jurafsky, and
  Yang}]{Pryzant2019AutomaticallyNS}
Reid Pryzant, Richard~Diehl Martinez, Nathan Dass, Sadao Kurohashi, Dan
  Jurafsky, and Diyi Yang. 2019.
\newblock \href {https://api.semanticscholar.org/CorpusID:208248333}
  {Automatically neutralizing subjective bias in text}.
\newblock \emph{ArXiv}, abs/1911.09709.

\bibitem[{Qian et~al.(2023)Qian, Han, He, Zheng, and
  Zheng}]{qian-etal-2023-sentiment}
Fan Qian, Jiqing Han, Yongjun He, Tieran Zheng, and Guibin Zheng. 2023.
\newblock \href {https://doi.org/10.18653/v1/2023.findings-acl.821} {Sentiment
  knowledge enhanced self-supervised learning for multimodal sentiment
  analysis}.
\newblock In \emph{Findings of the Association for Computational Linguistics:
  ACL 2023}, pages 12966--12978, Toronto, Canada. Association for Computational
  Linguistics.

\bibitem[{Ranaldi and Freitas(2024)}]{ranaldi-freitas-2024-aligning}
Leonardo Ranaldi and Andre Freitas. 2024.
\newblock \href {https://aclanthology.org/2024.eacl-long.109} {Aligning large
  and small language models via chain-of-thought reasoning}.
\newblock In \emph{Proceedings of the 18th Conference of the European Chapter
  of the Association for Computational Linguistics (Volume 1: Long Papers)},
  pages 1812--1827, St. Julian{'}s, Malta. Association for Computational
  Linguistics.

\bibitem[{Rodrigo-Ginés et~al.(2024)Rodrigo-Ginés, de~Albornoz, and
  Plaza}]{RODRIGOGINES2024121641}
Francisco-Javier Rodrigo-Ginés, Jorge~Carrillo de~Albornoz, and Laura Plaza.
  2024.
\newblock \href {https://doi.org/https://doi.org/10.1016/j.eswa.2023.121641} {A
  systematic review on media bias detection: What is media bias, how it is
  expressed, and how to detect it}.
\newblock \emph{Expert Systems with Applications}, 237:121641.

\bibitem[{R{\o}nningstad et~al.(2023)R{\o}nningstad, Velldal, and
  {\O}vrelid}]{ronningstad2023entity}
Egil R{\o}nningstad, Erik Velldal, and Lilja {\O}vrelid. 2023.
\newblock Entity-level sentiment analysis (elsa): An exploratory task survey.
\newblock \emph{arXiv preprint arXiv:2304.14241}.

\bibitem[{Samonte(2018)}]{10.1145/3230348.3230354}
Mary Jane~C. Samonte. 2018.
\newblock \href {https://doi.org/10.1145/3230348.3230354} {Polarity analysis of
  editorial articles towards fake news detection}.
\newblock In \emph{Proceedings of the 2018 1st International Conference on
  Internet and E-Business}, ICIEB '18, page 108–112, New York, NY, USA.
  Association for Computing Machinery.

\bibitem[{Tang et~al.(2023)Tang, Yang, Huang, Tam, and
  Tang}]{tang-etal-2023-finentity}
Yixuan Tang, Yi~Yang, Allen Huang, Andy Tam, and Justin Tang. 2023.
\newblock \href {https://doi.org/10.18653/v1/2023.emnlp-main.956}
  {{F}in{E}ntity: Entity-level sentiment classification for financial texts}.
\newblock In \emph{Proceedings of the 2023 Conference on Empirical Methods in
  Natural Language Processing}, pages 15465--15471, Singapore. Association for
  Computational Linguistics.

\bibitem[{Thoppilan et~al.(2022)Thoppilan, Freitas, Hall, Shazeer,
  Kulshreshtha, Cheng, Jin, Bos, Baker, Du, Li, Lee, Zheng, Ghafouri, Menegali,
  Huang, Krikun, Lepikhin, Qin, Chen, Xu, Chen, Roberts, Bosma, Zhou, Chang,
  Krivokon, Rusch, Pickett, Meier-Hellstern, Morris, Doshi, Santos, Duke,
  S{\o}raker, Zevenbergen, Prabhakaran, D{\'i}az, Hutchinson, Olson, Molina,
  Hoffman-John, Lee, Aroyo, Rajakumar, Butryna, Lamm, Kuzmina, Fenton, Cohen,
  Bernstein, Kurzweil, Aguera-Arcas, Cui, Croak, hsin Chi, and
  Le}]{Thoppilan2022LaMDALM}
Romal Thoppilan, Daniel~De Freitas, Jamie Hall, Noam~M. Shazeer, Apoorv
  Kulshreshtha, Heng-Tze Cheng, Alicia Jin, Taylor Bos, Leslie Baker, Yu~Du,
  Yaguang Li, Hongrae Lee, Huaixiu~Steven Zheng, Amin Ghafouri, Marcelo
  Menegali, Yanping Huang, Maxim Krikun, Dmitry Lepikhin, James Qin, Dehao
  Chen, Yuanzhong Xu, Zhifeng Chen, Adam Roberts, Maarten Bosma, Yanqi Zhou,
  Chung-Ching Chang, I.~A. Krivokon, Willard~James Rusch, Marc Pickett,
  Kathleen~S. Meier-Hellstern, Meredith~Ringel Morris, Tulsee Doshi,
  Renelito~Delos Santos, Toju Duke, Johnny~Hartz S{\o}raker, Ben Zevenbergen,
  Vinodkumar Prabhakaran, Mark D{\'i}az, Ben Hutchinson, Kristen Olson,
  Alejandra Molina, Erin Hoffman-John, Josh Lee, Lora Aroyo, Ravi Rajakumar,
  Alena Butryna, Matthew Lamm, V.~O. Kuzmina, Joseph Fenton, Aaron Cohen,
  Rachel Bernstein, Ray Kurzweil, Blaise Aguera-Arcas, Claire Cui,
  Marian~Rogers Croak, Ed~Huai hsin Chi, and Quoc Le. 2022.
\newblock \href {https://api.semanticscholar.org/CorpusID:246063428} {Lamda:
  Language models for dialog applications}.
\newblock \emph{ArXiv}, abs/2201.08239.

\bibitem[{Touvron et~al.(2023)Touvron, Martin, Stone, Albert, Almahairi,
  Babaei, Bashlykov, Batra, Bhargava, Bhosale, Bikel, Blecher, Ferrer, Chen,
  Cucurull, Esiobu, Fernandes, Fu, Fu, Fuller, Gao, Goswami, Goyal, Hartshorn,
  Hosseini, Hou, Inan, Kardas, Kerkez, Khabsa, Kloumann, Korenev, Koura,
  Lachaux, Lavril, Lee, Liskovich, Lu, Mao, Martinet, Mihaylov, Mishra,
  Molybog, Nie, Poulton, Reizenstein, Rungta, Saladi, Schelten, Silva, Smith,
  Subramanian, Tan, Tang, Taylor, Williams, Kuan, Xu, Yan, Zarov, Zhang, Fan,
  Kambadur, Narang, Rodriguez, Stojnic, Edunov, and Scialom}]{touvron2023llama}
Hugo Touvron, Louis Martin, Kevin Stone, Peter Albert, Amjad Almahairi, Yasmine
  Babaei, Nikolay Bashlykov, Soumya Batra, Prajjwal Bhargava, Shruti Bhosale,
  Dan Bikel, Lukas Blecher, Cristian~Canton Ferrer, Moya Chen, Guillem
  Cucurull, David Esiobu, Jude Fernandes, Jeremy Fu, Wenyin Fu, Brian Fuller,
  Cynthia Gao, Vedanuj Goswami, Naman Goyal, Anthony Hartshorn, Saghar
  Hosseini, Rui Hou, Hakan Inan, Marcin Kardas, Viktor Kerkez, Madian Khabsa,
  Isabel Kloumann, Artem Korenev, Punit~Singh Koura, Marie-Anne Lachaux,
  Thibaut Lavril, Jenya Lee, Diana Liskovich, Yinghai Lu, Yuning Mao, Xavier
  Martinet, Todor Mihaylov, Pushkar Mishra, Igor Molybog, Yixin Nie, Andrew
  Poulton, Jeremy Reizenstein, Rashi Rungta, Kalyan Saladi, Alan Schelten, Ruan
  Silva, Eric~Michael Smith, Ranjan Subramanian, Xiaoqing~Ellen Tan, Binh Tang,
  Ross Taylor, Adina Williams, Jian~Xiang Kuan, Puxin Xu, Zheng Yan, Iliyan
  Zarov, Yuchen Zhang, Angela Fan, Melanie Kambadur, Sharan Narang, Aurelien
  Rodriguez, Robert Stojnic, Sergey Edunov, and Thomas Scialom. 2023.
\newblock \href {http://arxiv.org/abs/2307.09288} {Llama 2: Open foundation and
  fine-tuned chat models}.

\bibitem[{Wang et~al.(2022)Wang, Wei, Schuurmans, Le, hsin Chi, and
  Zhou}]{Wang2022SelfConsistencyIC}
Xuezhi Wang, Jason Wei, Dale Schuurmans, Quoc Le, Ed~Huai hsin Chi, and Denny
  Zhou. 2022.
\newblock \href {https://api.semanticscholar.org/CorpusID:247595263}
  {Self-consistency improves chain of thought reasoning in language models}.
\newblock \emph{ArXiv}, abs/2203.11171.

\bibitem[{Wang et~al.(2023)Wang, Xie, Ding, Feng, and Xia}]{Wang2023IsCA}
Zengzhi Wang, Qiming Xie, Zixiang Ding, Yi~Feng, and Rui Xia. 2023.
\newblock \href {https://api.semanticscholar.org/CorpusID:258048703} {Is
  chatgpt a good sentiment analyzer? a preliminary study}.
\newblock \emph{ArXiv}, abs/2304.04339.

\bibitem[{Wei et~al.(2022)Wei, Wang, Schuurmans, Bosma, Xia, Chi, Le, Zhou
  et~al.}]{wei2022chain}
Jason Wei, Xuezhi Wang, Dale Schuurmans, Maarten Bosma, Fei Xia, Ed~Chi, Quoc~V
  Le, Denny Zhou, et~al. 2022.
\newblock Chain-of-thought prompting elicits reasoning in large language
  models.
\newblock \emph{Advances in neural information processing systems},
  35:24824--24837.

\bibitem[{Ye and Durrett(2022)}]{Ye2022TheUO}
Xi~Ye and Greg Durrett. 2022.
\newblock \href {https://api.semanticscholar.org/CorpusID:252873674} {The
  unreliability of explanations in few-shot prompting for textual reasoning}.
\newblock In \emph{Neural Information Processing Systems}.

\bibitem[{Zhang et~al.(2022)Zhang, Zhou, Chen, Bai, and
  He}]{zhang2022enhancing}
Qi~Zhang, Jie Zhou, Qin Chen, Qingchun Bai, and Liang He. 2022.
\newblock Enhancing event-level sentiment analysis with structured arguments.
\newblock In \emph{Proceedings of the 45th International ACM SIGIR Conference
  on Research and Development in Information Retrieval}, pages 1944--1949.

\bibitem[{Zhang et~al.(2023)Zhang, Deng, Liu, Pan, and
  Bing}]{zhang2023sentiment}
Wenxuan Zhang, Yue Deng, Bing Liu, Sinno~Jialin Pan, and Lidong Bing. 2023.
\newblock \href {http://arxiv.org/abs/2305.15005} {Sentiment analysis in the
  era of large language models: A reality check}.

\bibitem[{Zhong et~al.(2023)Zhong, Ding, Liu, Du, and Tao}]{Zhong2023CanCU}
Qihuang Zhong, Liang Ding, Juhua Liu, Bo~Du, and Dacheng Tao. 2023.
\newblock \href {https://api.semanticscholar.org/CorpusID:257050251} {Can
  chatgpt understand too? a comparative study on chatgpt and fine-tuned bert}.
\newblock \emph{ArXiv}, abs/2302.10198.

\bibitem[{Zhou et~al.(2021)Zhou, Wang, Zhang, and He}]{zhou-etal-2021-implicit}
Deyu Zhou, Jianan Wang, Linhai Zhang, and Yulan He. 2021.
\newblock \href {https://doi.org/10.18653/v1/2021.emnlp-main.551} {Implicit
  sentiment analysis with event-centered text representation}.
\newblock In \emph{Proceedings of the 2021 Conference on Empirical Methods in
  Natural Language Processing}, pages 6884--6893, Online and Punta Cana,
  Dominican Republic. Association for Computational Linguistics.

\end{thebibliography}

\bibliographystylelanguageresource{lrec-coling2024-natbib}
\bibliographylanguageresource{languageresource}

\end{document}